\documentclass[letterpaper, 10pt, conference]{ieeeconf}
\pdfoutput=1

\usepackage{graphicx}
\usepackage{subfigure}
\usepackage{amsmath,amssymb,amsopn,amstext,amsfonts}
\usepackage[space,nosort]{cite}
\usepackage{pdfsync}
\usepackage{mathtools}
\usepackage{algorithm}
\usepackage{algorithmic}
\usepackage{bm}
\usepackage{url}
\usepackage{float}
\usepackage{multirow}
\usepackage{enumerate}
\usepackage{array}

\IEEEoverridecommandlockouts
\title{\LARGE \bf Towards View-invariant and Accurate Loop Detection \\Based on Scene Graph}

\author{
Chuhao Liu and Shaojie Shen
\thanks{
All authors are from the Department of Electronic and Computer Engineering, Hong Kong University of Science and Technology, Hong Kong, China. {\tt\footnotesize cliuci@connect.ust.hk, eeshaojie@ust.hk}}%
}

\graphicspath{{fig/}}
\DeclareGraphicsExtensions{.pdf,.png,.jpg}

\begin{document}
\maketitle

\label{todolist}

\begin{abstract}
Loop detection plays a key role in visual Simultaneous Localization and Mapping (SLAM) by correcting the accumulated pose drift. In indoor scenarios, the richly distributed semantic landmarks are view-point invariant and hold strong descriptive power in loop detection. The current semantic-aided loop detection embeds the topology between semantic instances to search a loop. However, current semantic-aided loop detection methods face challenges in dealing with ambiguous semantic instances and drastic viewpoint differences, which are not fully addressed in the literature. This paper introduces a novel loop detection method based on an incrementally created scene graph, targeting the visual SLAM at indoor scenes. It jointly considers the macro-view topology, micro-view topology, and occupancy of semantic instances to find correct correspondences. Experiments using handheld RGB-D sequence show our method is able to accurately detect loops in drastically changed viewpoints. It maintains a high precision in observing objects with similar topology and appearance. Our method also demonstrates that it is robust in changed indoor scenes. 
\end{abstract}

\section{Introduction}

\begin{figure}[ht]
    \centering
    \subfigure[]{\includegraphics[width=0.75\columnwidth]{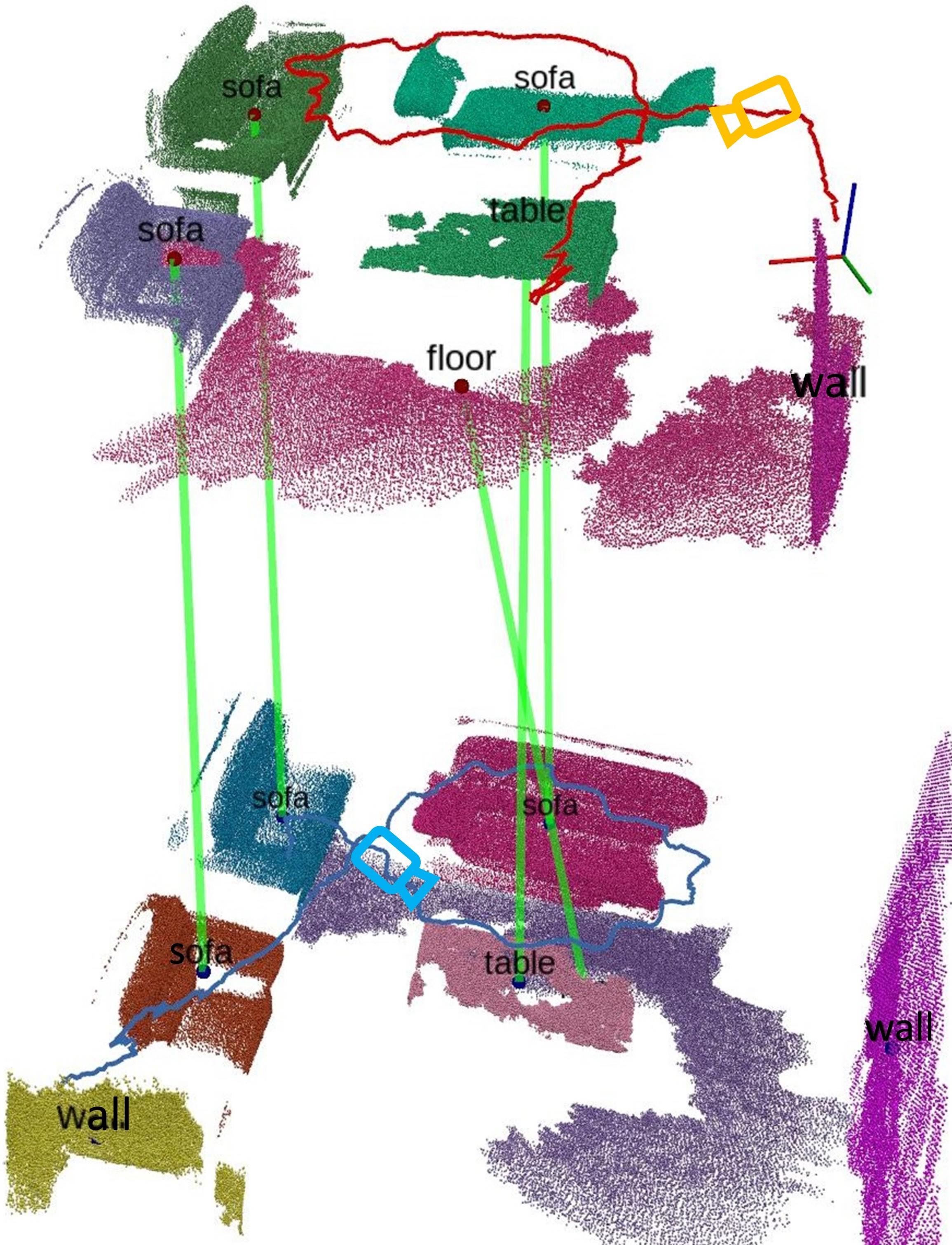}}
    \subfigure[]{\includegraphics[width=0.9\columnwidth]{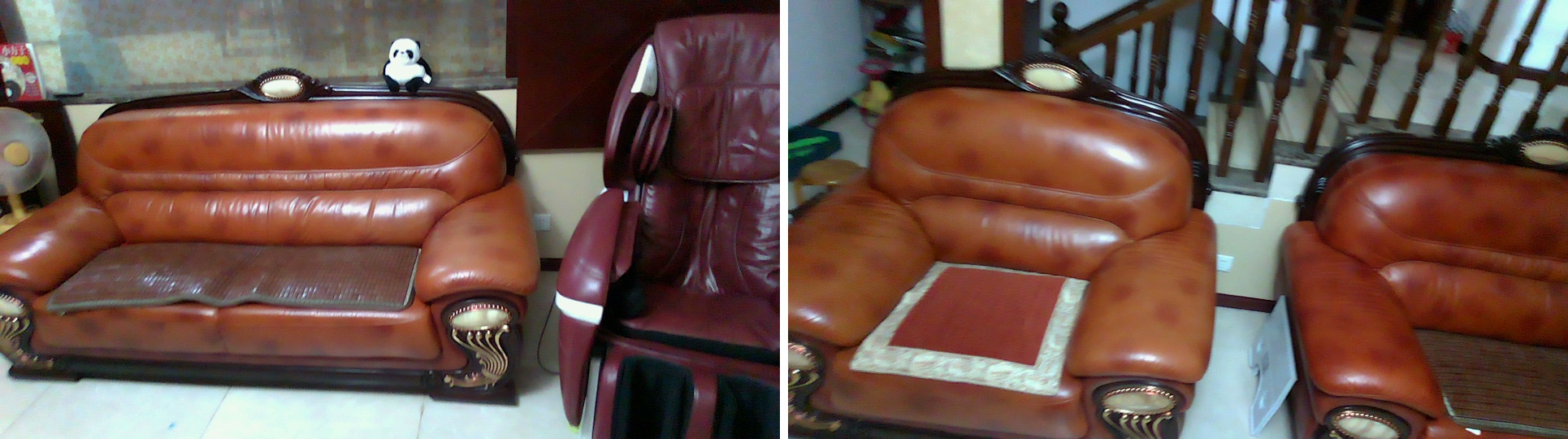}}
    \vspace{-0.3cm}	
    \caption{A dense \emph{living room} is scanned in two sessions with drastically changed viewpoints. (a) The active scene graph is shown in the upper half, while the inactive counterpart is in the bottom half. Their trajectories are colored in red and blue. The proposed method successfully detects the loop and finds the correct correspondences, which are visualized in green lines. (b) The sofas incorporate ambiguity to the data association that they are spatially close to each other and are in similar appearance.}
	\label{fig1}
	\vspace{-0.7cm}
\end{figure}

Loop detection is an important yet challenging task in visual SLAM. Current visual-inertial odometry (VIO) achieves promising results in local accuracy, while inevitably having accumulated pose drift in long term. Loop detection is able to correct the accumulated drift in VIO and re-localize the camera. A desired loop detection module should be accurate and invariant to viewpoint differences.

Appearance-based loop detection, such as DBoW2 \cite{bow2012galvez} and SuperGlue \cite{sarlin20superglue}, extracts image features and search correspondences between an image pair. It retrievals images from the database to match with the live image. SuperGlue demonstrates promising feature matching performance in their experiments. However, in scanning objects with similar texture or drastic viewpoint differences, appearance-based methods inevitably recall inaccurate loops or false loops.

On the other hand, indoor environments have rich types of semantic instances, which can be view-invariant landmarks in loop detection. Based on RGB-D camera, Fusion++ \cite{fusion++2018McCormac} utilizes instance segmentation and dense reconstruction to generate a 3D object map. The loop can be searched between instances based on the semantic type and instance appearance. Besides each instance's self attributes, inter-instance topology is discriminative in describing a semantic instance. Lin etc.\cite{Lin2021topobj} embeds instance topology through the random walk descriptor to match semantic graphs. Their method achieves loop detection in a sparse scene with significant viewpoint change.
However, this method struggles to handle ambiguous instances of the same semantic type that are located close to each other. Such instances are common in indoor scenes, as illustrated by the example of sofas shown in Fig. \ref{fig1}(a).
Because these instances share neighbors, simply embedding them using the random walk descriptor results in the same random walk route and identical topology description, which can lead to a false loop being recalled.

Beyond the ambiguous instances, arbitrary scanning trajectory incorporate variances to the generated semantic graph. When a dense scene is scanned from different viewpoints, only parts of the instances are co-observed. The reconstructed semantic graph is different due to small overlapped volumes. Another variance is the low-dynamic object, such as an opened curtain or a moved chair. They are frequently observed in the indoor scenes, and they create micro-differences in the reconstructed semantic graph. Utilizing the topology of semantic landmarks to detect loop should consider these variances and ensure robustness under variances.

Meanwhile, the performance of semantic-aided loop detection significantly depends on whether its front-end can construct persistent semantic landmarks. Previous methods use the object map \cite{fusion++2018McCormac}, \cite{Lin2021topobj} as their front end, which runs instance segmentation on images before the reconstruction.  Unfortunately, the instance segmentation on images is sensitive to viewpoint differences. Moreover, segmentation on image is variant to object shape, appearance, and occlusion, making the cross-frame data association extremely difficult. Thus, previous semantic-aided loop detection mostly do experiment at scenes with sparse objects or ignores many objects that cannot be associated confidently. 

\label{mypros}
As a result, relying on RGB-D camera, we propose a novel loop detection method by matching an active scene graph with an inactive scene graph in indoor environments.
Unlike previous methods using object maps as their front-end, our method constructs semantic landmarks based on SceneGraphFusion\cite{wu2021scenegraphfusion}, which segments a reconstructing 3D map into semantic instances in real-time. Since the semantic instances are extracted from an accumulated map, it does not require data association across frames.
In the back-end, we use a random walk descriptor to embed the global topology and a neighbor walk descriptor to embed the local topology of each instance. Along with an occupancy similarity, they are jointly considered in finding correspondences. 
The proposed method is evaluated in dense indoor environments with ambiguous instances and drastic viewpoint differences. The result shows our method outperforms the original random walk descriptor method with higher precision. In scanning objects with similar appearance, our method is more accurate than SuperGlue. At last, we evaluate our method in changed scenes and verifies our method is robust against micro-differences in the scene graphs.

\section{Related works}
\subsection{Semantic Mapping with RGB-D Camera}
In object mapping based on RGB-D camera, Fusion++ \cite{fusion++2018McCormac} proposes a novel fusion method that integrate the instance segmentation of Mask R-CNN \cite{he2017mask} into an object map. Voxblox++ \cite{voxblox++2019Grinvald} segments depth map and fuse them into a global geometric segments. Results of Mask R-CNN are integrated into geometric segments. However, a common downside of their method is that the Mask R-CNN is sensitive to viewpoints. The results of instance segmentation contain a large variance. So, the cross-frame data association can easily fail and extracted semantic landmarks are of poor quality. Moreover, persistent instances, such as cabinets, beds, and walls, are seldom detected due to partial observation. Similarly, Kimera \cite{Kimera2020Rosinol} uses semantic segmentation on the image and propagates the semantic label into a 3D global mesh map. Based on the metric-semantic meshes, Kimera generates a hierarchical scene graph with multiple layers. 

In scene segmentation, semantic instances are segmented from a reconstructed 3D model \cite{han2020occuseg},\cite{Chen2021hais}. SceneGraphFusion \cite{wu2021scenegraphfusion} incrementally segments an reconstructing scene in real-time RGB-D SLAM. It first generates a geometric segment map. Then, it proposes a graphic neural network (GNN) to predict the semantic label of each segment node and inter-segment relationship. The scene graph is incrementally constructed, including semantic instances.

Both object mapping and scene segmentation can construct a semantic map. The key difference is segmentation before the reconstruction or segmentation after the reconstruction. Compared with object mapping methods, SceneGraphFusion is less sensitive to the viewpoint variances and able to reconstruct the persistent instances such as sofa and cabinet.

\subsection{Semantic-aided Loop Detection}

Given the commonly existed IMU sensor in RGB-D camera, the camera pose suffers from drift in x, y, z, and yaw angle \cite{qin2018relocalization}, which is 4 Degree-of-Freedom (DoF). Beyond the viewpoint invariance, the loop detection method should be invariant to 4-DoF pose drift. 


The semantic objects are discriminative landmarks in loop detection. The similarity of a potential correspondence can be calculated relies on object appearance \cite{fusion++2018McCormac}, \cite{Qian2022covloop} and orientation \cite{Li2020orientobj}. 
Based on Kimera \cite{Kimera2020Rosinol}, Hydra \cite{hughes2022hydra} embeds a hierarchical descriptor with layers: room, place, and appearance descriptor. Hence, the loop can be searched in a coarse-to-fine strategy. It significantly improves loop detection efficiency. However, their common downside is that the appearance correspondences are hard constraints to recall a loop. And viewpoint invariance cannot be improved.

On the other hand, the topology of semantic landmarks holds high descriptive power.
The random walk descriptor is firstly used in X-View \cite{Abel2018xview} to detect loop.  
For each vertex in the reconstructed semantic graph, X-View conducts random walks along connected vertices and records the semantic label of the visited vertices. It successfully embeds each vertex topology and used it to compute similarity in the graph matching stage. The random walk descriptor is also successfully implemented in later works \cite{Lin2021topobj}, \cite{Liu2019gloc}.
However, the random walk descriptor can not handle ambiguous instances and is sensitive to variances in semantic graphs.

Attributes of the random walk descriptor can be explained through the property proved in node2vec\cite{grover2016node2vec}. 
Before the random walk descriptor is applied in SLAM, it is successfully used in large graph embedding \cite{grover2016node2vec}, \cite{perozzi2014deepwalk}, such as social networks. As illustrated in node2vec, there are two walking strategies in the random walk descriptor, breadth-first sampling (BFS) and depth-first sampling (DFS). The former represents a micro-view of the graph, while the latter represents a macro-view. The X-View proposes the random walk descriptor with a purely DFS strategy. The DFS embeds vertex topology based on homophily, which represents the vertex co-appearances in the graph. Thus, ambiguous semantic instances with shared neighbors are likely to be falsely associated.
Moreover, the DFS strategy is more sensitive to graph variance, especially when the vertex is described in inexact features. As a result, the random walk descriptor is degenerated in variances caused by viewpoints or segmentation.

To ensure discriminative description for instance topology and improve robustness in the drastic view-point differences, we jointly use random walk descriptor and neighbor walk descriptor to embed the instance topology. The former represents its macro-view topology in the graph, while the latter represents its micro-view topology.

\section{Methodology}
\vspace{-0.2cm}
\begin{figure}[ht]
	\includegraphics[width=0.95\columnwidth]{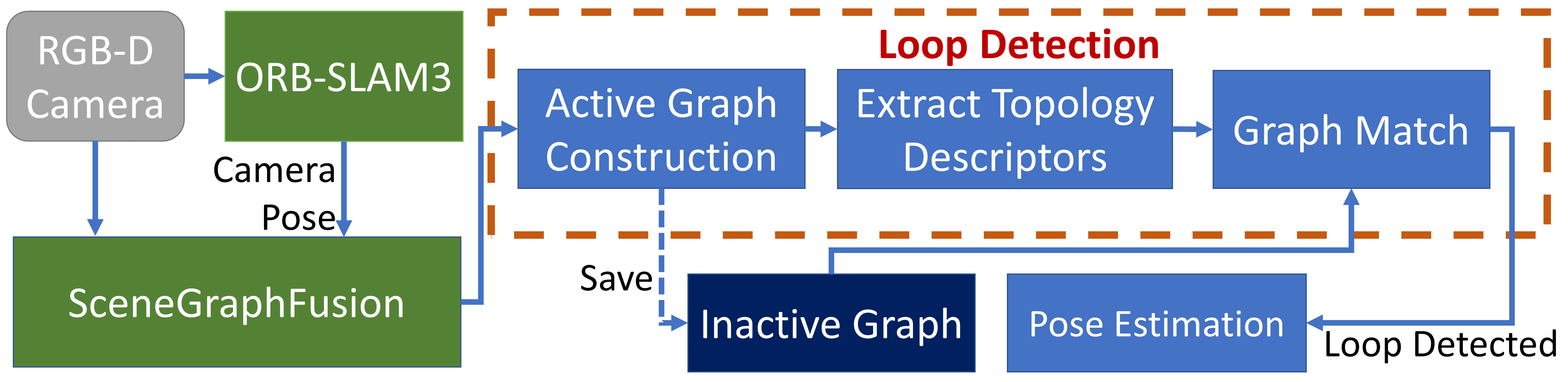}
	\caption{System structure}
	\label{fig_sys}
\end{figure}

\subsection{System Overview}
Our system takes RGB-D images as input and compute camera pose using ORB-SLAM3 \cite{campos2021orb3}. SceneGraphFusion \cite{wu2021scenegraphfusion} at the front end incrementally segments the reconstructed scene. Based on its result, an active scene graph is constructed and topology descriptors of each instance are created. Then, we match the active graph with a previously built inactive graph to detect loops. If a loop is recalled, pose drift between the graphs is estimated, and the scene graphs are fused.

\subsection{Active Scene Graph Construction}
In the first step, we construct a scene graph $\mathcal{G={\{V,E\}}}$ based on SceneGraphFusion \cite{wu2021scenegraphfusion}, where $\mathcal{V}$ is a vertex set while $\mathcal{E}$ is an edge set. 

To generate $\mathcal{V}$, we take the nodes from SceneGraphFusion as input. Some nodes can be over-segmented parts of object instances. We merge those nodes that are connected by an edge with the "same part" label as an instance. Those instances that are too small or predicted with low confidence are ignored. Each filtered instance is created as a vertex $v_i=\{\mathbf{p,b,n},l\},v_i \in \mathcal{V}$, with centroid $\mathbf{p} \in \mathbb{R}^3$, axis-aligned bounding box size $\mathbf{b} \in \mathbb{R}^3$, normal vector $\mathbf{n} \in \mathbb{R}^3$ and semantic label $l$. The normal vector is only used for surface types instances: wall and floor. Since the pose drift is 4-DoF, vertices have their roll and pitch orientation aligned with each other.

In constructing edge set $\mathcal{E}$, we add directed edges considering the semantic and geometric attributes of vertices. For two objects, an edge is added between them if their centroids are closer than a threshold. For an object $v_i$ and a vertical wall $v_j$, 
an edge is added if their perpendicular distance is smaller than a threshold. For two walls, they are connected if their centroids are close and the angle of their normal vector is larger than a threshold. For any type of vertex and a floor, they are connected if they are horizontally overlapped.  The fundamental principle of edge construction is embedding those discriminative topological relationships of each instance.

The real-time constructing scene graph is an active scene graph $\mathcal{G}^a$. It can be saved and loaded as an inactive graph $\mathcal{G}^i$ in the re-scan. 

\subsection{Topology Descriptors}
\label{sec_descriptor}
\begin{figure}[ht]
	\centering
	\includegraphics[width=0.88\columnwidth]{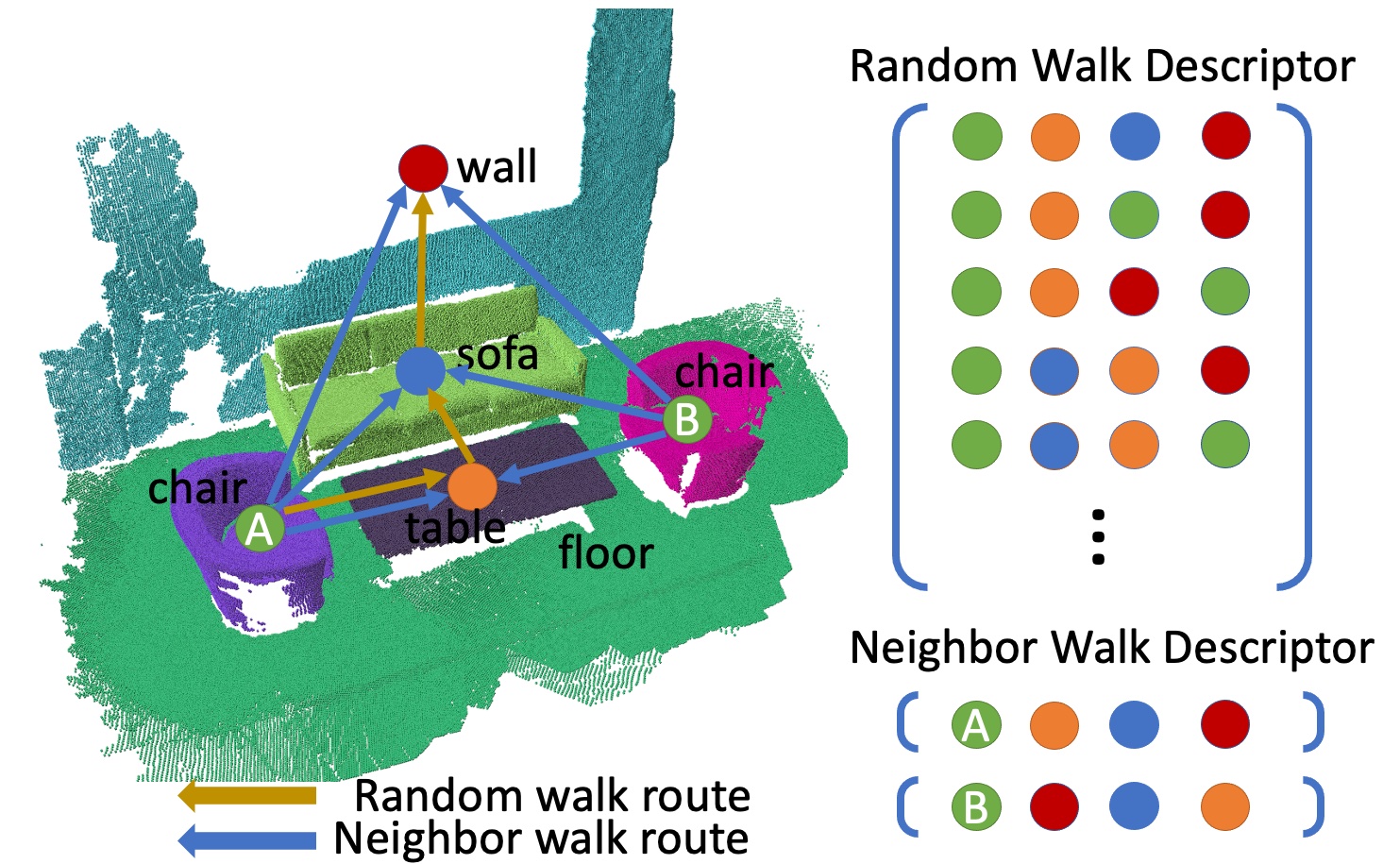}
	\vspace{-0.3cm}
	\caption{An example of topology descriptors. Each instance is marked with a centroid colored by its semantic type. The two chairs are ambiguous instances. The random walk descriptor on the top right can be used to describe any of the chairs, while their neighbor walk descriptors at the bottom right are embedded differently.}
	\label{fig_descriptors}
	\vspace{-0.3cm}
\end{figure}

In constructing $\mathcal{G}^a$, the topology descriptors of each vertex are incrementally generated as well.
To describe the macro-view topology of vertex $v_i$, we extract its random walk descriptor as $\mathbf{d^r_i} \in C^{N_i \times k}$ where $N_i$ is the total number of random walks and the walk depth $k=4$. In each walking step, it records the semantic type of the visited nodes and generates the $\mathbf{d^r_i}$. Similar to X-View, strategies that prevent walking back to visited nodes are applied. The random walk descriptor embeds a vertex based on homophily. However, it cannot distinguish ambiguous instances, such as chairs shown in Fig. \ref{fig_descriptors}. 

To handle ambiguous instances and improve robustness, a neighbor walk descriptor $\mathbf{d^n_i} \in C^{M_i \times q}$ is extracted for vertex $v_i$, where $M_i$ is the total number of neighbor walks and walk depth $q=4$. 
We walk through the neighbors of $v_i$ in anticlockwise order along the z-axis. If the angle between the current neighbor and the last neighbor is larger than $150^{\circ}$, the walk is rejected. The semantic label of visited neighbors are recorded and finally embedded as $\mathbf{d^n_i} $.
As shown in Fig. \ref{fig_descriptors}, the neighbor walk descriptor embeds the ambiguous chairs differently. 
The neighbor walk descriptor is inspired by the BFS strategy in node2vec \cite{grover2016node2vec}. It highlights the structural equivalence of vertex. And it becomes a descriptive feature to each vertex. Another strength of the neighbor walk descriptor is its invariance to 4-DoF pose drift. The local topology subgraph is consistent in 4-DoF pose drift.

In creating both of the topology descriptors, walking route over a floor vertex is rejected. Otherwise, nearly every vertex connects with a floor vertex and degenerate the discriminative of a descriptor. The completed topology descriptors of vertex $v_i$ are embedded as $\{\mathbf{d^r_i,d^n_i}\}$. 

\subsection{Graph Match}
Graph match is conducted between $\mathcal{G}^a$ and $\mathcal{G}^i$ to find correct correspondences between them. 

\subsubsection{Build Score Matrix}
In this step, the similarity score of all possible pairs of nodes is computed. For a specific pair of nodes $\{v_i,v_j|v_i \in \mathcal{V}^{a}, v_j \in \mathcal{V}^{i}\}$, their similarity score is calculated:
\vspace{-0.2cm}
\begin{equation}
	\label{equ_score}
	\mathcal{S}_{i,j} =\sigma_r(\mathbf{d^r_i,d^r_j}) +\lambda_n \cdot \sigma_n(\mathbf{d_i^n,d_j^n})+\lambda_v \cdot \sigma_{v}(\mathbf{b_i,b_j}).
\end{equation}
\vspace{-0.6cm}

The random walk term $\sigma_r(\mathbf{d^r_i,d^r_j})$ and the neighbor walk term $\sigma_n(\mathbf{d^n_i,d^n_j})$ can be computed by counting the number of identical rows between descriptors. They are normalized to $[0,1]$.

\vspace{-0.4cm}
\begin{equation*}
	\sigma_r(\mathbf{d^r_i,d^r_j}) = \frac{c(\mathbf{d_i^r,d_j^r})}{\min(|\mathbf{d_i^r}|,|\mathbf{d_j^r}|)}, \\
	\sigma_n(\mathbf{d^n_i,d^n_j}) = \frac{c(\mathbf{d_i^n,d_j^n})}{\max(|\mathbf{d^n_i}|,|\mathbf{d^n_j}|)}
\end{equation*}
\vspace{-0.1cm}
The $c(\mathbf{d_i,d_j})$ is the number of identical rows between descriptors $\mathbf{d_i}$ and $\mathbf{d_j}$. And $|\mathbf{d_i}|$ is the total number of rows of the descriptor.

Notice that the random walk term is normalized over the minimum number of descriptors. If some of the random walks in $v_i$ are missed due to an incomplete scan, the random walk term can still generate a high similarity score once most of $\mathbf{d^r_i}$ can find an identical descriptor in $\mathbf{d^r_j}$. On the other hand, the neighbor walk term is normalized over the maximum number of descriptors. If $v_i$ has inexact neighbor walks as $v_j$ while some of its neighbor walks can find identical counterparts in $v_j$, it generates a low similarity score. 
Hence, it sets a high boundary for matching two neighbor walk descriptors and rejects potential ambiguous loop pairs.

To further improve loop accuracy, the volume similarity term is calculated: 
\begin{equation}
	\sigma_v(\mathbf{b_i,b_j}) ={\frac{l(\mathbf{b_i})-l(\mathbf{b_j})}{\max(l(\mathbf{b_i}),l(\mathbf{b_j}))}}
\end{equation}
where $\mathbf{b_i} \in \mathbb{R}^3$ is the axis-aligned bounding box of instance $i$ and $l(\mathbf{b_i})$ is its diagonal length.
Then, similarity score $S_{i,j}$ is calculated and normalized to $[0,1]$. A complete score matrix $\mathcal{S} \in \mathbb{R}^{|\mathcal{V}^s|\times|\mathcal{V}^t|}$ is built. The parameter $\lambda_n,\lambda_v$ are term's weights that can be adjusted.

\subsubsection{Find Correspondences}
We search the correspondences based on the score matrix $\mathcal{S}$. The max similarity for each row and column in $\mathcal{S}$ is extracted, while the other elements are all set to zero. If the maximum similarity is larger than a threshold $\tau =0.5$, we mark it as a pair of matched instances. Unlike the X-View, our method allows an active node to match with only one inactive node. 

Since the wall is frequently reconstructed in indoor scenes, they are more likely to be matched falsely. We set a verification for the matched pairs of walls:
they are verified correctly if at least one of their neighbors is also matched accordingly.
Finally, if a minimum number of correspondences $\epsilon = 4$ is found, a loop is recalled. 

\vspace{-0.2cm}
\subsection{Register Graphs}
Once a loop is recalled, the matched instances in $\mathcal{G}^a$ are constructed as a set of point cloud $\mathcal{P}^a$. Similarly, an inactive point cloud $\mathcal{P}^i$ is generated. $\{\mathcal{P}^a,\mathcal{P}^i\}$ are aligned using the dense registration method \cite{zhou2016fastreg} and their relative pose $T^a_i$ is calculated.

\vspace{-0.2cm}
\section{Experiment}
\begin{figure*}[ht]
	\includegraphics[width=\textwidth]{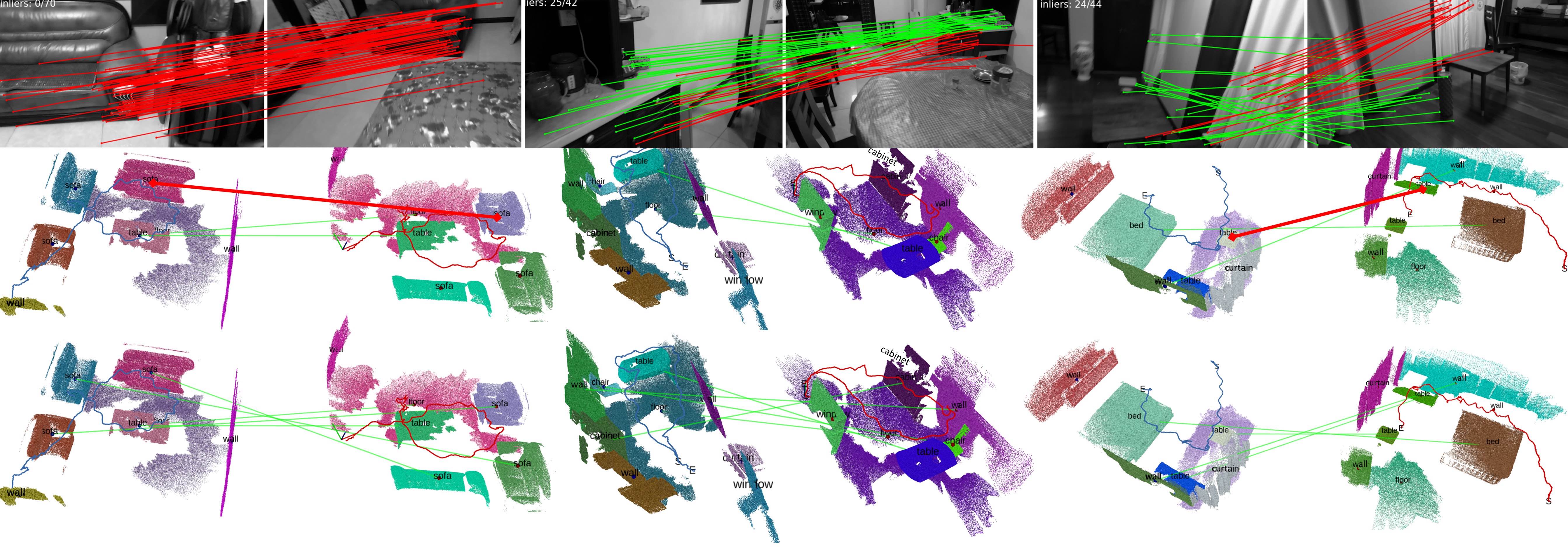}
	\vspace{-0.9cm}
	\caption{ The loop results in the self collected RGB-D dataset at scenes: \emph{living room}, \emph{dining room} and \emph{bedroom}, are shown from left to right. The first row is the result of SuperGlue. The second row is a result of the random walk descriptor method. The last row is a result of the proposed method. In scene graphs and matched images, correct correspondences are colored in green, while outliers are in red.}
	\label{fig_result0}
	\vspace{-0.9cm}
\end{figure*}

The evaluation is conducted in two real-world experiments. Firstly, we collect RGB-D camera data in 4 dense and static indoor scenes. In each scene, we move the camera in two sessions, whose scanning sequence is opposite to each other. It generates keyframe with drastic viewpoint differences, where we evaluate the loop detection performance. 
Secondly, we evaluate the system in indoor scenes with micro-differences. We select two groups of sequences from the 3RScan dataset \cite{wald2019rio} and one group of self-collected sequences. Each scene partially changes between the two session's scans. 

The customized RGB-D dataset is collected using the Intel Realsense L515 camera. Its RGB image is $960 \times 540$ dimension, while depth is $640 \times 480$. Camera pose is generated from ORB-SLAM3 \cite{campos2021orb3}. In the experiments with the 3RScan sequences, we use the provided camera pose. All the experiments are run on a computer with an Intel i7 CPU @ 2.8GHz and an Nvidia GTX1070 GPU.

In the proposed method, parameters $\lambda_r= 1.0, \lambda_n= 0.5, \lambda_v = 0.6$. Considering the neighbor walk term $\lambda_n$ sets a high boundary in accepting a loop and the axis-aligned bounding box gives a volume term $\lambda_v$ with variance, these two terms are set to lower values. 
In applying the SuperGlue \footnote{https://github.com/magicleap/SuperGluePretrainedNetwork}, RGB images at the original resolution is used. We sample $20\%$ of the images from the active scan and  the inactive scan. Each sampled active image is queried against all the sampled inactive images. 

In each pair of scene graphs, we label those pairs of instances belonging to correct matches. If an instance is over-segmented, we label the match between all the broken instances to the target instance as a correct match. Three evaluation metrics are used: Recall ratio is the ratio of matched frames over the total number of frames; Precision is the ratio of correctly matched instances over the total number of matches; Matching score is the number of matched instances over the total number of active instances. 

\subsection{Scans in Drastic Viewpoint Variations}

\vspace{-0.4cm}
\begin{table}[ht]
	\centering
	\begin{tabular}{c|c c c}
	\hline
	{Method} & {RWD} & {NWD} & {Proposed} \\
	\hline
	\hline
	{Recall ($\%$)} & 13.9 & 7.0 & $\mathbf{28.4}$ \\
	\hline
	{Precision ($\%$)} & 58.9 & $\mathbf{100.0}$ & $\mathbf{100.0}$ \\
	\hline
	{Match Score ($\%$)} & 42.9 & 14.3 & $\mathbf{70.3}$ \\
	\hline
	\end{tabular}
	\caption{Loop detection results on the collected RGB-D dataset.}
	\vspace{-0.8cm}
	\label{tab1}
\end{table}

Our method is compared with the purely random walk descriptor (RWD) and neighbor walk descriptor (NWD) methods. To ensure the comparison is fair, all the methods use identical module to construct scene graphs. As shown in Table. \ref{tab1}, our method achieves the highest recall rate and precision. In the scenes with ambiguous instances, false correspondences are generated in the random walk descriptor method, such as the \emph{bedroom} and \emph{living room} shown in Fig. \ref{fig_result0}. In the \emph{dinning room} scene, the random walk descriptor misses some of the correct correspondences due to the small overlapped volume between the two scans. 
Meanwhile, the purely neighbor walk descriptor is able to precisely find correspondences, thanks to its discriminative local topology information with spatial awareness. But it only recalls a small number of loops due to unequally observed instances between the two scans. 
On the other hand, our method jointly uses random walk descriptor, neighbor walk descriptor and volume term. It can distinguish the ambiguous instances and is robust in variances caused by arbitrary scanning trajectories. The results show our method achieves precise loop detection and a higher recall rate. 

Since the object-based loop detection approaches \cite{Lin2021topobj}, \cite{Qian2022covloop} are not open sourced, we cannot directly compare with them. However, as the scenes shown in figure \ref{fig_result0}, our experiment environment is dense and involves ambiguous nearby objects. And we evaluate the method in drastic viewpoint variations. These factors pose serious challenges to loop detection which are seldom considered in previous related works.

\vspace{-0.3cm}
\label{result_superglue}
\begin{table}[ht]
	\centering
	\begin{tabular}{c|c c}
		\hline
		& SuperGlue &Proposed \\
		\hline
		\hline
		Recall $(\%)$ & \textbf{69.93}  & 28.4  \\
		\hline
		Instance Precision& N/A & \textbf{100.0} \\
		\hline
		Features Precision & 62.03 & N/A \\
		\hline
		Runtime & 5350.28 ms  &\textbf{48.26 ms}  \\
		\hline

	\end{tabular}
	\caption{Comparison between SuperGlue and the proposed method.}
	\label{tab_superglue}
	\vspace{-0.6cm}
\end{table}

Another baseline to compare with is SuperGlue \cite{sarlin20superglue}.
Here, a pair of associated features is labelled as a false positive if its epipolar distance is larger than a threshold. And precision is the ratio between the number of true positive correspondences and the total number of correspondences. In calculating epipolar errors, we uses the poses computed from ORB-SLAM3 and align the poses between the two scans by densely register their fully reconstructed map. Because the camera poses contains uncertainty, we set a large threshold for the epipolar distance at $1$ pixel. This evaluation focus on analyzing the obviously false feature correspondences created by SuperGlue. 

As shown in TABLE \ref{tab_superglue}, SuperGlue detects loops at higher recall rate than our method. However, it creates plenty of outliers in drastic viewpoint differences, especially in sub-volumes with similar appearance or repeated textures. In the \emph{living room} as shown in Fig. \ref{fig_result0}, false loops are called when observing two sofas in similar appearance. Actually, outliers in sub-volumes with similar appearance are also shown in the indoor pose estimation experiment in SuperGlue \cite{sarlin20superglue}. This is a natural limitation of the appearance-based method. The result drives us to think appearance-based method and semantic-aided method can be jointly used in loop detection task to improve recall rate while maintaining a high precision. 

In runtime analysis, we summarize the frame-wise runtime, including SceneGraphFusion's execution time. As shown in Table \ref{tab_superglue}, SuperGlue is extremely time-consuming. It has complexity at $O(|\mathcal{I}^i|)$, where $\mathcal{I}^i$ is the set of previous images. The complexity grows unbounded along scanning duration. On the other hand, our method and the related semantic-aided loop detection methods run much faster, thanks to the sparsity of semantic graphs. More importantly, our method has complexity at $O(\mathcal{|V|}^a*\mathcal{|V|}^i)$. It grows with explored volumes, rather than the scanning duration.

\vspace{-0.1cm}
\subsection{Micro-differences in the scene}

\vspace{-0.3cm}
\begin{table}[h]
	\centering
	\begin{tabular}{c |c c c |c c c}
		\hline
		 \multirow{2}{*}{}& \multicolumn{3}{c|}{RWD} & \multicolumn{3}{c}{Proposed Method} \\
		\cline{2-7}
		 & Rec. & Pre. &MS  & Rec. & Prc. &MS\\
		\hline
		\hline
		\emph{3RScan-Office} & 91.7& 94.3 &29.4 &91.7 & \textbf{99.9} & 36.4
		 \\
		\hline
		\emph{3RScan-LR} &35.5 & 80.4 & 33.1 &37.6 & \textbf{100.0} & 38.6
		 \\
		\hline
		\emph{Studyroom} & 20.2 & \textbf{100.0} & 40.0 &20.2 & \textbf{100.0} & 40.0
		\\
		\hline
	\end{tabular}
	\caption{Loop detection in changed scenes. Metrics in $\%$ 
	}
	\label{tab_dynamic}
	\vspace{-0.5cm}
\end{table}

\begin{figure*}[ht]
	\centering		 
    \includegraphics[width=0.9\textwidth]{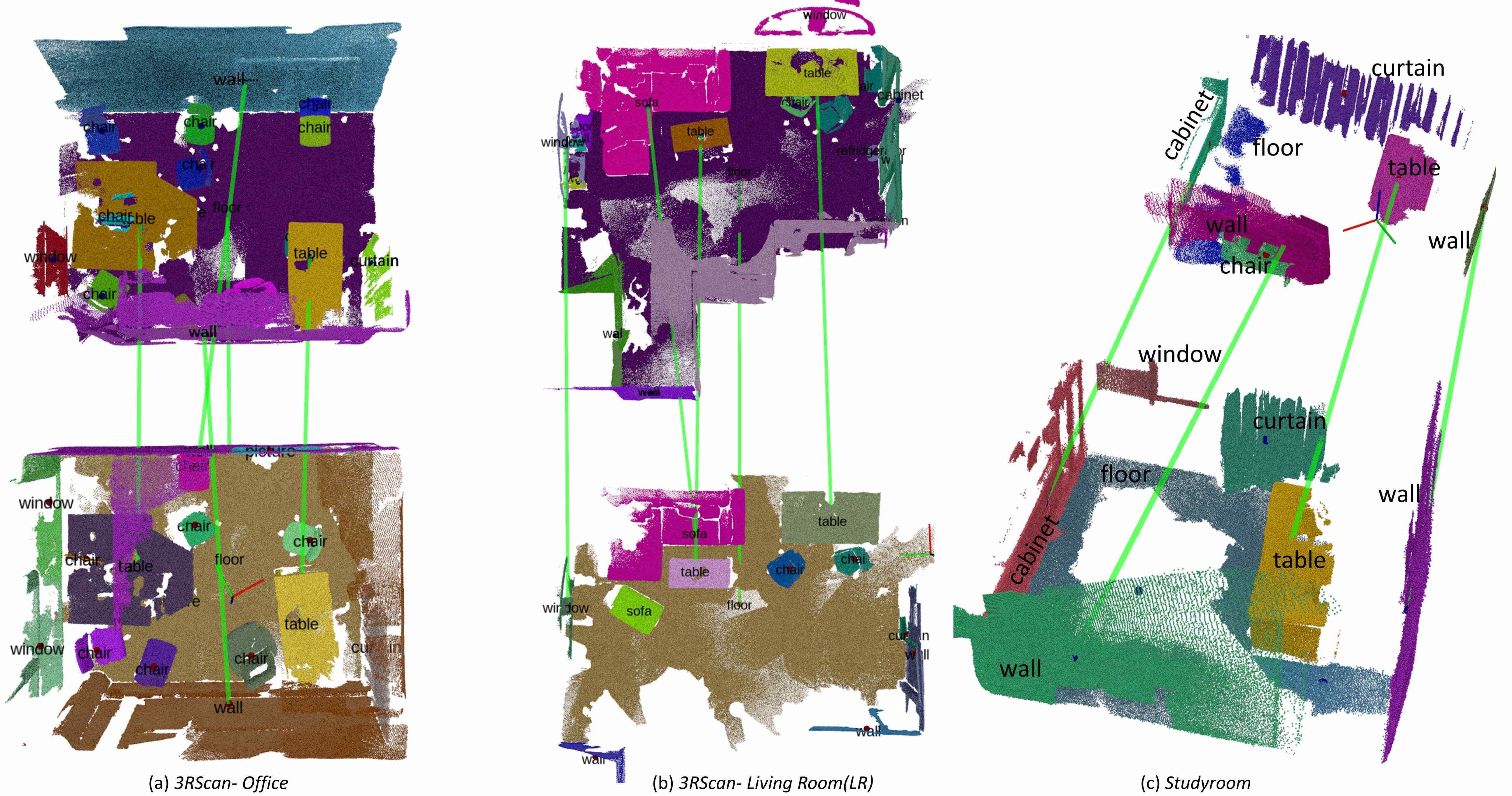}
	\vspace{-0.2cm}
	\caption{The proposed methods in three changed scenes. (a) Multiple chairs are moved; (b) Table and chairs are moved. A sofa set is split;  (c) Curtain is opened. All the matched pairs of instances are colored in green lines.}
	\vspace{-0.9cm}
	\label{fig_dynamic}
\end{figure*}

As shown in Fig. \ref{fig_dynamic}, indoor scene change in a re-scan due to moving objects or an opened curtain. We evaluate the robustness of our method in variances caused by dynamic objects. We evaluate the loop accuracy of those persistent semantic landmarks in a scene. In all the scenes, dynamic objects, such as chairs and curtains, are not considered as correspondences. However, dynamic objects are used in constructing topology descriptors.

As shown in Table. \ref{tab_dynamic} and Fig. \ref{fig_dynamic}, our method detects loops precisely in the changed scenes. 
Since these scenes do not have many ambiguous instances, the purely random walk descriptor achieves higher precision than the last experiment. However, our method still outperforms the purely random walk descriptor in precision.

\vspace{-0.2cm}
\begin{table}[h]
	\centering
	\begin{tabular}{c|c c c}
		\hline
		  & Recall & Precision &Match Score \\
		\hline
		\hline
		  \emph{3RScan-Office} & 56.6 & \textbf{96.7} & 24.3 \\
		\hline
		  \emph{3RScan-LR} & 35.5 & 66.2 & 13.9 \\
		\hline
		  \emph{Studyroom} &  0.0 & N/A & N/A \\
		\hline
	\end{tabular}
	\caption{The proposed method in the filtered graph.}
	\label{tab_filter}
	\vspace{-0.8cm}
\end{table}

To further evaluate the influence of dynamic objects, we remove them from building topology descriptors and generate a filtered graph. As shown in Table \ref{tab_filter}, performance in all the scenes is affected. In the \emph{3RScan-Living Room}, the neighbor walk cannot distinguish the ambiguous tables, without the topology connecting nearby chairs. In the \emph{studyroom}, which is a sparse scene, some vertices cannot build any valid neighbor walk descriptor once the curtain is removed. So, it fails to detect any loop in the filtered graph. Based on the comparison, we can see low-dynamic objects play an important role in accurate loop detection. Even though some of them are ambiguous to be matched, they enhance the descriptive of their nearby instances.And the descriptive is only slightly affected by objects movement.

It is worth noticing that the SceneGraphFusion inevitably creates false labeled instances or over-segmentation. This is a common challenge for scene segmentation tasks. In both of the experiments, our method is robust in variances caused by inaccurate scene segmentation. On the other hand, with the development of scene segmentation which provides more accurate semantic landmarks, loop detection should achieve a higher recall rate and matching score. 

\vspace{-0.1cm}
\subsection{Pose estimation}
\vspace{-0.4cm}
\begin{figure}[H]
	\centering
	\includegraphics[width=0.7\columnwidth]{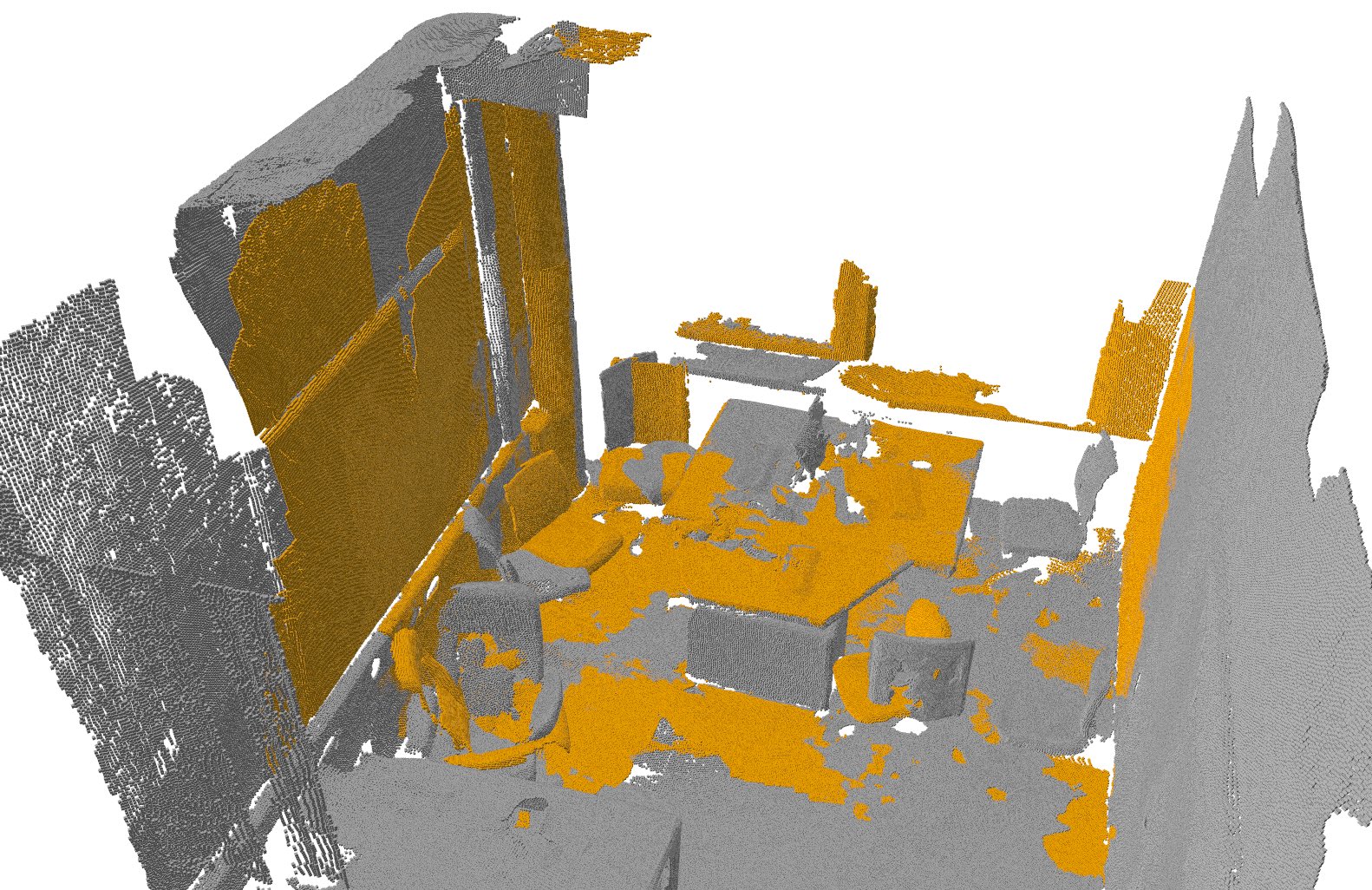}
	\vspace{-0.3cm}
	\caption{The aligned scene graphs in \emph{3RScan-Office}.}
	\label{fig_register}
	\vspace{-0.7cm}
\end{figure}

\begin{table}[h]
	\centering
	\begin{tabular}{c|c c}
		\hline
		     & Rotation Error & Translation Error \\
		\hline
	    3RScan Scenes & $0.047^\circ$ & 0.132 m \\
		\hline
	\end{tabular}
	\caption{Averaged pose error at \emph{3RScan- Office and LR}.}
	\label{tab_register}
	\vspace{-0.8cm}
\end{table}

Once a loop is detected, the matched instances are densely registered \cite{zhou2016fastreg} and the relative pose $T^a_i$ is calculated. We evaluate the pose estimation accuracy in 3RScan scenes using its ground truth pose. The pose accuracy is shown in Table \ref{tab_register}, while one of the qualitative results is shown in Fig. \ref{fig_register}.

\section{Conclusion}
To sum up, we propose a novel loop detection method based on scene graphs. The macro-view topology, micro-view topology, and occupied volume of each semantic landmark are jointly considered in finding correspondences. We evaluate our performance dense indoor scenes with drastic viewpoint variance. The result shows our method outperforms the random walk descriptor in recall rate and precision. Compared with SuperGlue, our method is accurate in observing sub-volumes with similar appearance. In variances caused by micro-differences in the scenes, our method demonstrates strong robustness.


\bibliography{lch}

\end{document}